\documentclass[11pt]{article}    
\usepackage[a4paper, total={6.5in, 10in}]{geometry}
\usepackage[utf8]{inputenc}
\usepackage[colorlinks,citecolor=blue,urlcolor=magenta,bookmarks=false,hypertexnames=true]{hyperref}

\usepackage[round]{natbib}
\renewcommand\cite{\citep}
\usepackage{booktabs}
\usepackage{graphicx}
\usepackage{multirow}
\usepackage{adjustbox}
\usepackage{CJKutf8}
\usepackage{subcaption}
\usepackage{booktabs}
\usepackage{array}
\usepackage{natbib}
\usepackage{hyperref}

\title{Global Beats, Local Tongue: Studying Code Switching in K-pop Hits on Billboard Charts}

\author{
	Aditya Narayan Sankaran, Reza Farahbakhsh, Noel Crespi\\
	SAMOVAR, Télécom SudParis\\
	Institut Polytechnique de Paris\\
    91120 Palaiseau, France\\
	\href{mailto:aditya.sankaran@ip-paris.fr}
	{\texttt{\small{aditya.sankaran@ip-paris.fr}}}
}
\date{}

\newcounter{RZNumberOfComments}
\stepcounter{RZNumberOfComments}

\begin{document}
\maketitle
\begin{abstract}
    Code switching, particularly between Korean and English, has become a defining feature of modern K-pop, reflecting both aesthetic choices and global market strategies. This paper is a primary investigation into the linguistic strategies employed in K-pop songs that achieve global chart success, with a focus on the role of code-switching and English lyric usage. A dataset of K-pop songs that appeared on the Billboard Hot 100 and Global 200 charts from 2017 to 2025, spanning 14 groups and 8 solo artists, was compiled. Using this dataset, the proportion of English and Korean lyrics, the frequency of code-switching, and other stylistic features were analysed. It was found that English dominates the linguistic landscape of globally charting K-pop songs, with both male and female performers exhibiting high degrees of code-switching and English usage. Statistical tests indicated no significant gender-based differences, although female solo artists tend to favour English more consistently. A classification task was also performed to predict performer gender from lyrics, achieving macro F1 scores up to 0.76 using multilingual embeddings and handcrafted features. Finally, differences between songs charting on the Hot 100 versus the Global 200 were examined, suggesting that, while there is no significant gender difference in English, higher English usage may be more critical for success in the US-focused Hot 100. The findings highlight how linguistic choices in K-pop lyrics are shaped by global market pressures and reveal stylistic patterns that reflect performer identity and chart context.
\end{abstract}

\section{Introduction} 
    \begin{CJK}{UTF8}{mj}
    Korean Pop (K-pop) has seen a rise in popularity in the western music circles, with K-pop groups selling out stadium shows similar to their famous western counterparts~\cite{saeji2024making, liu2020branding}. K-pop music has a distinct sound that is similar to Western pop, but leans more towards an experimental direction, incorporating more than two genres in a single song. As described by many media outlets, K-pop songs tend to be visually striking with colourful and vibrant music videos, which have enabled them to achieve true global commercial success and gain mass appeal across multiple countries and languages. With the super hit song \textit{"Gangnam Style"} by PSY, streaming sites have seen a rise in K-pop's presence in mainstream Western music charts, with big players like BTS and BLACKPINK dominating the Billboard Hot 100 and Global 200 charts for weeks on end~\cite{liu2020branding, caulfield2022blackpink}. 
        
    \begin{figure}[ht]
        \centering
        \includegraphics[width=0.7\linewidth]{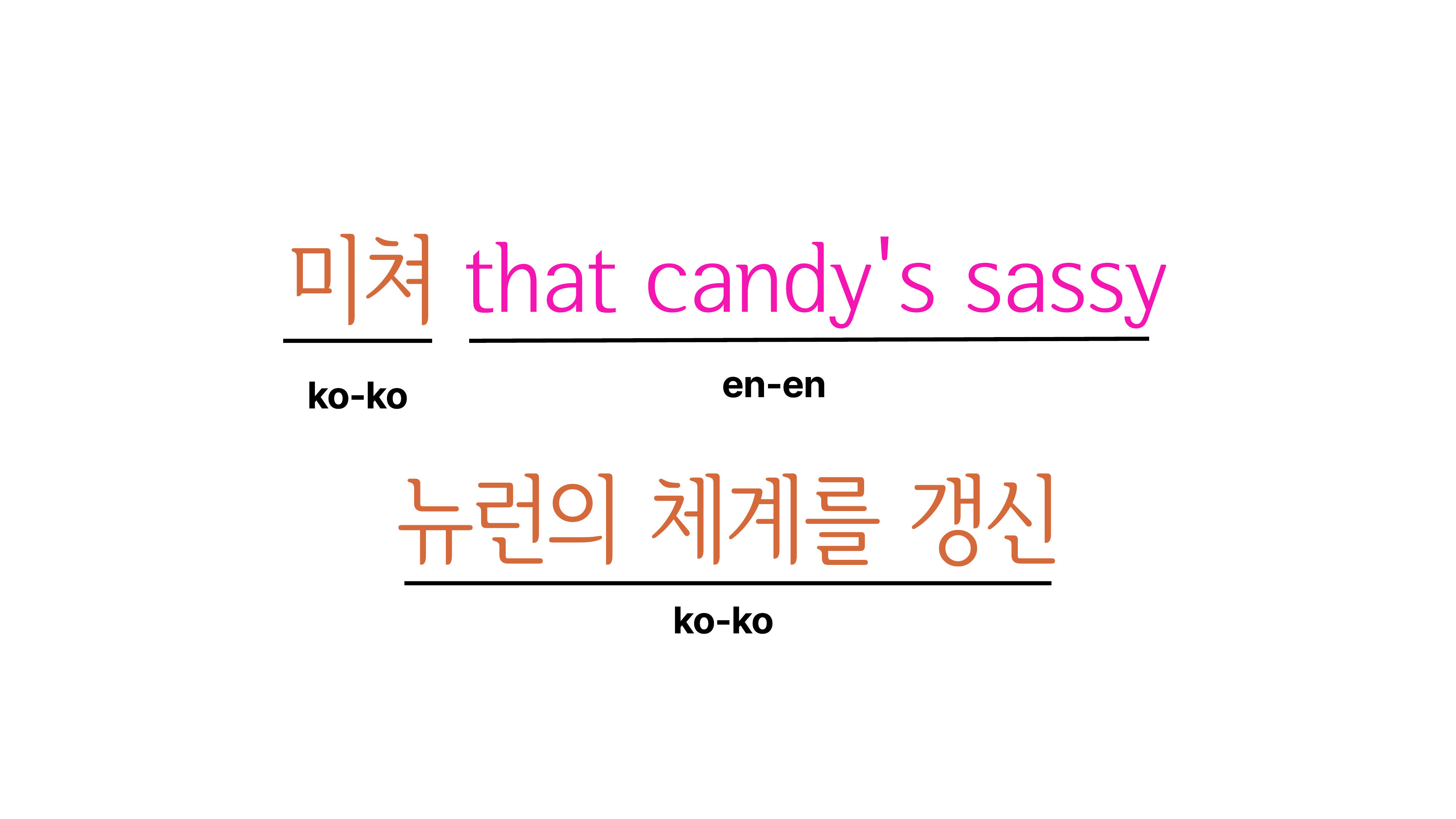}
        \caption{A snippet from LE SSERAFIM's song Crazy, where the presence of Code Switching is clear with 2 languages (ko-Korean, en-English). The first half of the under-text is the language, and the second half is the script.}
        \label{fig:intro-ex-1}
    \end{figure}
    
    While K-pop, as its name suggests, is primarily in Korean, we have seen a new wave of songs that feature a mix of English and Korean Lyrics. The recent 2024 blockbuster song titled "APT" by Rosé and Bruno Mars, which repeatedly uses the word 'APT’ as a hook, is a romanised abbreviation of the Korean term “아파트”, which is pronounced “apateu” in English, meaning ‘apartment’.
    \end{CJK} As K-pop songs have become increasingly popular in the Western world, the industry has also been adapting to this shift, incorporating more English into its songs. This mixing of English and Korean lyrics brings up an interesting concept from Linguistics called Code-switching. Code-switching (CS) refers to the mixing, by bilinguals (or multilinguals), of two or more languages in discourse, often with no change of interlocutor or topic~\cite{poplack2001code}. Scholars note that as K-pop’s fanbase expanded globally, the use of English in lyrics became increasingly common, evolving from a taboo-breaking tool into an index of “fun” and global connection~\cite{Schneider2023EnglishsEL}. An excerpt from the song CRAZY by LESSERAFIM, as shown in figure~\ref{fig:intro-ex-1}, displays this very idea. As shown, there is a mix of languages, ranging from English to Korean.
    
    Despite widespread acknowledgement of English code-mixing in K-pop~\cite{sriautarawong2025investigating, berliana2022such, margaretta2024korean}, there is a gap in understanding how this practice might differ by artist demographics and whether it correlates with chart performance. In particular, this study asks: Do male (boy group) and female (girl group) K-pop acts differ in their use of English vs. Korean in lyrics when aiming for international charts? Additionally, we explore whether solo artists show a similar pattern and how the degree of Korean/English lyrics relates to chart success (e.g. appearing in top chart positions). Addressing these questions can shed light on whether certain groups rely more on English lyrics as a strategy for global popularity, and whether lyrical content carries distinctive features that differentiate, say, a boy group’s song from a girl group’s song.
    
    In this work, we conduct a quantitative analysis of K-pop lyrics from songs that charted on Billboard’s Hot-100 and Global-200 lists between Jan 2017 and May 2025. This work also presents a novel perspective on a computational linguistics approach to code-switching in music. We construct a unique dataset that combines chart performances of K-pop artists with their song lyrics, and conduct two main investigations:
    \begin{enumerate}
        \item A text classification task to predict whether a song is by a boy group or a girl group based on its lyrics 
        \item A statistical comparison of linguistic features (English/Korean usage and code-switching frequency) between different artist categories (male vs. female, group vs. solo) in the Billboard charts.
    \end{enumerate}


    Our experiments revealed meaningful patterns – for instance, we find that virtually all internationally successful K-pop songs contain a high proportion of English (often well above 50\% of the lyrics), and while male and female groups both heavily code-switch, male acts tend to include slightly more Korean in their lyrics on average. However, these gender differences are not large enough to be statistically significant in our sample. We also demonstrate that a simple model can distinguish male-group vs. female-group lyrics with about 70\% accuracy, suggesting there are subtle but learnable differences in word choice and language use. 

\section{Related Works}

    Over the past decades, code-switching has become a major research focus in NLP, evolving from linguistics-based studies to advanced machine learning approaches. Recent surveys highlight the development of benchmarks, models, and tasks specifically designed for code-switched data, such as language identification, part-of-speech tagging, sentiment analysis, and natural language inference. Multilingual models, especially those fine-tuned on code-switched data, have shown the best performance, but challenges remain due to the complexity and variability of code-switching patterns across languages and contexts~\cite{Winata2022The, Khanuja2020GLUECoS, Cetinoglu2016Challenges, Bansal2020CodeSwitching}. While NLP methods have been widely adapted for music information retrieval (MIR) and symbolic music generation—treating music as a sequence of symbols similar to language—there is limited direct research on code-switching in music lyrics within the NLP community. Most MIR research leverages NLP tools for tasks like author detection, content generation, and music classification, but the specific phenomenon of code-switching in lyrics (such as in K-pop) is not yet a major focus in computational music studies~\cite{Le2024Natural}. 
    
    Recent research has focused on the types and functions of code-switching in K-pop lyrics, particularly the use of English alongside Korean, and its potential impact on international popularity, including success on the Billboard charts. Studies analysing songs by groups such as NewJeans, ASTRO, and Red Velvet have identified various forms of code-switching (e.g., inter-sentential, intra-sentential, tag-switching) and highlighted the importance of accurate pronunciation and the strategic use of English for audience engagement and global recognition~\cite{margaretta2024korean, berliana2022such, Jocelin2019Code, Schneider2023EnglishsEL}. While some research suggests that code-switching can enhance a song's appeal among bilingual and international audiences, the relationship between linguistic hybridity and chart success is complex, influenced by factors such as promotional support, pronunciation, and the unique functions of English usage~\cite{margaretta2024korean, berliana2022such, Schneider2023EnglishsEL}. Broader studies on code-switching in Asian pop music and advertising further contextualise these findings, suggesting that language mixing is often a deliberate strategy to adapt to global markets and negotiate cultural identity~\cite{Picone2024Lyrical, Ahn2017Language, Pu2010On}.

    Despite substantial progress in modelling code-switching within natural language processing (NLP)—including tasks such as language identification, sentiment analysis, and natural language inference—most research has focused on conversational or social media text. While multilingual models fine-tuned on code-switched data have shown promise, the phenomenon remains underexplored in creative domains like music. Although recent sociolinguistic studies have examined the functions and types of code-switching in K-pop lyrics and their potential impact on global visibility, there remains little computational work that rigorously analyses how such language use correlates with measurable success on international charts like the Billboard Hot 100 or Global 200. This paper addresses that gap by applying NLP methods to a novel curated dataset of Billboard-charting K-pop songs to quantify English–Korean code-switching patterns, examine their distribution across artist categories, and evaluate their relationship with chart performance, thereby bringing computational insight into a phenomenon largely studied through qualitative lenses.
    
\section{Dataset}

    Our dataset consists of K-pop songs (both group and solo artists) that appeared on the Billboard Hot 100 (US singles chart) or Billboard Global 200 chart between January 2017 and May 2025. We focused on K-pop acts who have made a significant international chart impact, as presented in Table~\ref{tab:kpop-stars}. In total, we gathered data for 22 artists – specifically, 11 groups (7 female groups and 4 male groups) and 10 solo artists (4 female and 6 soloists). These include globally renowned groups like BTS, BLACKPINK, TWICE, Stray Kids, etc., as well as soloists such as Lisa and Rose (members of BLACKPINK), J-Hope and V (members of BTS), among others. For each artist, we compiled all songs that entered either the Hot 100 or the Global 200 during the period of interest.

    \begin{table}[h]
        \centering
        \begin{subtable}[t]{0.45\textwidth}
            \centering
            \begin{tabular}{@{}p{4cm}cc@{}}
                \toprule
                \textbf{Group Artist} & \textbf{Gender} & \textbf{Category} \\
                \midrule
                aespa & F & G \\
                BLACKPINK & F & G \\
                ITZY & F & G \\
                LE SSERAFIM & F & G \\
                NewJeans & F & G \\
                Red Velvet & F & G \\
                TWICE & F & G \\
                BTS & M & G \\
                ENHYPEN & M & G \\
                SEVENTEEN & M & G \\
                Stray Kids & M & G \\
                \bottomrule
            \end{tabular}
            \caption{K-pop Group Artists by Gender}
        \end{subtable}
        \hfill
        \begin{subtable}[t]{0.45\textwidth}
            \centering
            \begin{tabular}{@{}p{4cm}cc@{}}
                \toprule
                \textbf{Solo Artist} & \textbf{Gender} & \textbf{Category} \\
                \midrule
                Jennie & F & S \\
                Jisoo & F & S \\
                Lisa & F & S \\
                Rosé & F & S \\
                J-Hope & M & S \\
                Jimin & M & S \\
                Jin & M & S \\
                Jung Kook & M & S \\
                RM & M & S \\
                V & M & S \\
                \phantom{X} & \phantom{X} & \phantom{X}\\
                \bottomrule
            \end{tabular}
            \caption{K-pop Solo Artists by Gender}
        \end{subtable}
        \caption{K-pop Artists Included in the Dataset}
        \label{tab:kpop-stars}
    \end{table}

    \subsection{Data Collection}
    
    \paragraph{Artist Metadata}: Using the list of K-pop groups and soloists to extract their entire discography and the lyrics from Genius, which is an eponymous website that serves as a database for song lyrics, news stories, sources, poetry, and documents, in which users can provide annotations and interpretations. This list, extracted using \texttt{lyricgenius~\footnote{\url{https://github.com/johnwmillr/LyricsGenius}}}, would serve as a master list of all the songs for each of the artists. 
    \paragraph{Billboard Chart Data}: Our work collected data from 2 different sources for 2 different tasks. The ranking and related data were collected from Billboard Charts, with a focus on the Hot 100 and Global 200, which are the most popular charts. \texttt{billboard-charts~\footnote{\url{https://pypi.org/project/billboard.py/}}} was used for extracting billboard data. We concentrated on the above-mentioned period of time and collected data every Friday, as new music data is typically released on that day.
    
    The features collected from the Billboard chart include the following: the \texttt{title} of the song, the \texttt{artist} or artists involved, the song’s weekly \texttt{rank}, its highest position on the chart recorded as \texttt{peakPos}, the \texttt{lastPos} indicating its position in the previous week, the total number of weeks the song has appeared on the chart given by \texttt{weeks}, and a binary indicator \texttt{isNew} which shows whether the song is newly entered into the chart.

    \paragraph{Final Dataset}: A master folder dataset~\footnote{The datasets and accompanying code can be found at \url{https://github.com/callmesanfornow/k-pop-billboard.git}} was compiled, using the billboard data and genius data, with the following columns for each song: \texttt{Title}, \texttt{Artist}, \texttt{Peak Position} (on the respective chart), \texttt{Weeks on Chart}, \texttt{Chart Trajectory} (week-by-week positions as a time series), \texttt{Artist Gender} (M/F), \texttt{Artist Category} (Group or Solo), \texttt{Lyrics} (original), and \texttt{Cleaned Lyrics}, separated by hot100 and global200. The lyrics were cleaned by removing annotations and standardising spacing and punctuation. The cleaned lyrics were used for all analyses.
    
    \subsection{Dataset Statistics}

    \begin{table}[htbp]
        \centering
        \scriptsize
        \begin{subtable}[t]{0.48\textwidth}
            \centering
            \caption{Groups}
            \begin{adjustbox}{max width=\textwidth}
            \begin{tabular}{@{}p{3cm} c p{3.8cm} c@{}}
            \toprule
            \textbf{Artist} & \textbf{Gender} & \textbf{Charts} & \textbf{Total Songs} \\
            \midrule
            aespa & F & Global 200 & 11 \\
            BLACKPINK & F & Hot 100, Global 200 & 24 \\
            ITZY & F & Global 200 & 3 \\
            LE SSERAFIM & F & Hot 100, Global 200 & 9 \\
            NewJeans & F & Hot 100, Global 200 & 19 \\
            Red Velvet & F & Global 200 & 2 \\
            TWICE & F & Hot 100, Global 200 & 11 \\
            BTS & M & Hot 100, Global 200 & 29 \\
            ENHYPEN & M & Global 200 & 3 \\
            SEVENTEEN & M & Global 200 & 6 \\
            Stray Kids & M & Hot 100, Global 200 & 9 \\
            \bottomrule
            \end{tabular}
            \end{adjustbox}
        \end{subtable}
        \hfill
        \begin{subtable}[t]{0.48\textwidth}
            \centering
            \caption{Solo Artists}
            \begin{adjustbox}{max width=\textwidth}
            \begin{tabular}{@{}p{3cm} c p{3.8cm} c@{}}
            \toprule
            \textbf{Artist} & \textbf{Gender} & \textbf{Charts} & \textbf{Total Songs} \\
            \midrule
            Jennie & F & Hot 100, Global 200 & 14 \\
            Jisoo & F & Global 200 & 2 \\
            Lisa & F & Hot 100, Global 200 & 11 \\
            Rosé & F & Hot 100, Global 200 & 13 \\
            J-Hope & M & Hot 100, Global 200 & 8 \\
            Jimin & M & Hot 100, Global 200 & 12 \\
            Jin & M & Hot 100, Global 200 & 6 \\
            Jung Kook & M & Hot 100, Global 200 & 14 \\
            RM & M & Global 200 & 3 \\
            V & M & Hot 100, Global 200 & 15 \\
            \phantom{Blank} & \phantom{X} & \phantom{Blank} & \phantom{X} \\
            \bottomrule
            \end{tabular}
            \end{adjustbox}
        \end{subtable}
        \caption{K-pop Groups and Solo Artists by Gender, Chart Presence, and Total Songs}

        \label{tab:dataset-stats}
    \end{table}
        
    As presented in Table~\ref{tab:dataset-stats}, the dataset consists of K-pop groups and solo artists, categorised by gender, chart presence (Hot 100 and/or Global 200), and total number of songs. It includes 11 groups (7 female, 4 male) and 10 solo artists (4 female, 6 male). Group entries are led by BTS (29 songs) among males and BLACKPINK (24 songs) among females. Among soloists, V (15 songs) and Jung Kook (14 songs) are the most prolific male artists, while Jennie leads the female soloists with 14 songs. All artists appear on the \textit{Global 200} chart, with most also charting on the \textit{Hot 100}. The dataset captures the gender distribution, chart reach, and number of song across major K-pop acts.
    
    \section{Methodologies}

    With the dataset, we perform two tasks to analyse how code-switching is used in the context of Western audiences since Billboard charts are primarily Western. Also, an analysis of what makes a lyric suitable for a boy group or a girl group. Furthermore, we investigate the role of code-switching in the context of boy groups and girl groups, examining whether they code-switch more to achieve popularity on the Billboard charts.

    \subsection{Feature Extraction}

    In addition to processing the raw lyrics text, several quantitative features were engineered to capture underlying linguistic patterns present in the songs. The \textbf{Korean Lyric Ratio} was calculated as the proportion of words written in Hangul script, offering insight into the extent of Korean language usage. Complementarily, the \textbf{English Lyric Ratio} measured the fraction of words composed in the Latin alphabet, reflecting the prevalence of English in the lyrics. These were extracted using regular expressions, with whitespace being the delimiter to tokenise. To analyse multilingual blending, the \textbf{Code-Switch Count} quantified the number of language alternations at the word level. Each transition between Korean and English within a lyrical line was identified and counted, capturing the dynamic nature of code-switching in the dataset. Repetitive lyrical structures were assessed through the \textbf{Repetition Count}, which detected consecutive repeated n-grams (up to five words), encompassing patterns such as recurring chorus lines or expressive phrases like ``la la la.'' Phonetic stylistic choices were examined using the \textbf{Alliteration Count}, focusing primarily on English lyrics. Instances of consecutive words sharing the same initial sound, such as in the phrase ``bring the boys back'', were identified to quantify this rhetorical device. Lastly, the \textbf{Filler Words Count} tallied occurrences of common interjections and non-lexical expressions (e.g., ``oh,'' ``yeah,'' ``la''), which frequently appear in K-pop lyrics irrespective of the primary language. Collectively, these features provide a structured and interpretable linguistic profile of multilingual and stylistic patterns in K-pop songwriting.

    All features above were computed for each song in our dataset and are presented in Table~\ref{tab:tab-stats}. This enriched dataset enabled both a classification task and statistical analyses, as described next. Across both charts, songs performed by female artists consistently had a higher proportion of English lyrics and filler words, as well as more frequent use of alliteration, repetition, and code-switching compared to male artists. Male artists tended to have a higher ratio of Korean lyrics. Notably, the Hot 100 chart exhibited greater language and stylistic contrasts by gender, with female artists using English more extensively (approximately 89\%) and more filler words ($\approx$35 counts) compared to their male counterparts (75\% English, 26 filler counts).
    
    \begin{table}[h!]
    \centering
    \begin{adjustbox}{max width=\textwidth}
    \begin{tabular}{llccccccc}
    \hline
    \textbf{Chart} & \textbf{Gender} & \textbf{Kor Ratio} & \textbf{Eng Ratio} & \textbf{Filler Count} & \textbf{Alliteration} & \textbf{Repetition} & \textbf{Code Switches} \\
    \hline
    \multirow{2}{*}{Global 200} & F & 0.171 & 0.827 & 31.736 & 20.165 & 6.275 & 25.319 \\
                                & M & 0.238 & 0.755 & 24.632 & 13.235 & 4.412 & 20.574 \\
    \hline
    \multirow{2}{*}{Hot 100} & F & 0.112 & 0.888 & 35.107 & 24.036 & 7.429 & 16.964 \\
                             & M & 0.236 & 0.752 & 26.000 & 14.486 & 5.838 & 19.054 \\
    \hline
    \end{tabular}
    \end{adjustbox}
    \caption{Mean scores for heuristic linguistic features by chart and gender}
    \label{tab:tab-stats}
    \end{table}
    
    \subsection{Lyrics Classification}
    
    The primary investigation is to assess whether the lyrical content of a K-pop song can signal the gender of the performing artist—male or female—through a binary text classification task. This includes both groups and solo acts. The objective is to determine whether systematic linguistic or stylistic differences exist between male- and female-led performances. Each song was represented using a combination of dense multilingual text embeddings and handcrafted linguistic features. Specifically, \verb|XLM-RoBERTa|~\cite{conneau2019unsupervised} and \verb|GTE-multilingual|~\cite{zhang2024mgte} were used to encode lyrics into 768-dimensional vectors, taking the mean of the last hidden state across tokens. These embeddings were then reduced to 30 dimensions via Principal Component Analysis (PCA) to retain key variance. They were then concatenated with heuristic features such as English/Korean word ratios, code-switch counts, repetitions, filler word usage, and alliteration, yielding a hybrid feature set that captures both semantic content and surface-level stylistic traits.

    Using this representation, multiple classifiers were trained (Logistic Regression (LR), Support Vector Machine (SVM), Random Forest (RF), XGBoost, a Multilayer Perceptron (MLP), and a voting ensemble) to predict artist gender. To examine whether different linguistic patterns emerge depending on chart context, we treated Billboard Hot 100 and Global 200 songs as distinct classification tasks. This separation allowed us to explore whether lyrical cues are more predictive of gender in U.S.-charting songs versus globally-charting ones and to assess how classification performance varies with market orientation.
    
    Our aim is not to draw normative or stereotypical conclusions about gender in K-pop, but to treat male and female-led songs as distinct experimental groups for linguistic analysis. There are measurable stylistic and lexical patterns that exist across gendered performances. Rather than framing these as rigid differences, we interpret them as characterisations of lyrical style shaped by artistic, commercial, or cultural factors.

     \subsection{Statsistical Hypothesis Testing}

     The second part of our methodology tests specific hypotheses about code-switching frequency and language proportion differences between categories of artists.

     We formulated several comparisons:

    \noindent\textbf{Gender difference in English usage:} Do boy groups and girl groups differ in how much English (versus Korean) they use in their lyrics? Our primary metric for this was the average Korean lyric ratio (with the English ratio being complementary). We hypothesised that female groups—who often target the pop market—might use more English, whereas male groups might retain more Korean lyrics.
    
    \noindent\textbf{Gender difference in code-switch count:} Do male and female groups differ in how frequently they switch languages within a song? For instance, one might wonder if one group tends to alternate languages more line-by-line, while another might use a large English section all at once.
    
    \noindent\textbf{Solo vs. group differences:} We also compare solo female and solo male artists on these metrics. There is anecdotal evidence~\cite{Garza2021Where, Auh2024Creativity} that female soloists often release English-heavy songs (e.g., BLACKPINK members’ solos) targeting Western audiences, whereas male soloists—often rappers—might retain more Korean. 
    
    \noindent\textbf{Chart context:} Finally, we examine whether chart context influences language use. Specifically, we compare Hot 100 songs to Global 200 songs within the same category. For example, do girl group songs that enter the U.S. Hot 100 contain less Korean (i.e., more English) than girl group songs that chart only on the Global 200 (which includes Asian markets) and similarly for boy groups. This analysis helps indicate whether artists adapt their language use based on the target audience, such as U.S. listeners in the case of the Hot 100.
    
     For each group-wise comparison, a consistent statistical procedure was applied to test differences in linguistic features such as Korean lyric ratio and code-switching frequency. We calculated group-wise means and used Welch’s t-test to assess significance without assuming equal variances. Since most distributions were non-normal (Shapiro-Wilk p < 0.001), Mann-Whitney U test was used to compare medians. We report all test statistics and p-values, using a significance threshold of $\alpha$ = 0.05. Results with p < 0.05 are considered significant; borderline cases are noted cautiously. This analysis complements our classification experiments by isolating which features differ across gender or chart category.
     
    \section{Experiments}
    
    
    \subsection{Classification Experiment Setup}

    We trained and evaluated classification models on two subsets of data: (1) a smaller set of songs that appeared on the Billboard Hot 100, consisting mostly of high-profile hits (heavily dominated by artists such as BTS and BLACKPINK), and (2) a larger set of songs from the Billboard Global 200, which includes some Hot 100 entries along with additional globally popular tracks. For each subset, we performed an 80/20 stratified train-test split, preserving the gender ratio (male vs. female songs). The training set was used to fit each model—Logistic Regression (LR), Support Vector Machine (SVM), Random Forest (RF), XGBoost, Multi-Layer Perceptron (MLP), and a voting ensemble—while the held-out test set was used for evaluation. Given the relatively small dataset size, models were trained with default hyperparameters without extensive tuning. We report classification performance using both accuracy and macro-averaged F1-score on the test sets.

    \subsection{Statistical Test Experiment Setup}
    
    For each hypothesis—gender, solo artist, and chart differences—we filtered and labelled the data accordingly. For group-level gender analysis, we compared boy-group (Group 0) and girl-group (Group 1) songs, excluding solos. Solo comparisons followed the same labelling for male and female soloists. Chart-based analysis compared Hot 100 vs. Global 200 songs within each gender group. We computed code-switch counts and Korean lyric ratios, assessed normality using the Shapiro-Wilk test, and applied Welch’s t-test or Mann-Whitney U as appropriate. Given skewed distributions, non-parametric tests were generally more reliable. To account for limited sample sizes, we report both p-values and group means to illustrate potential trends even when not statistically significant.

    \section{Results and Discussions}

    \begin{figure}
        \centering
        \includegraphics[width=0.85\linewidth]{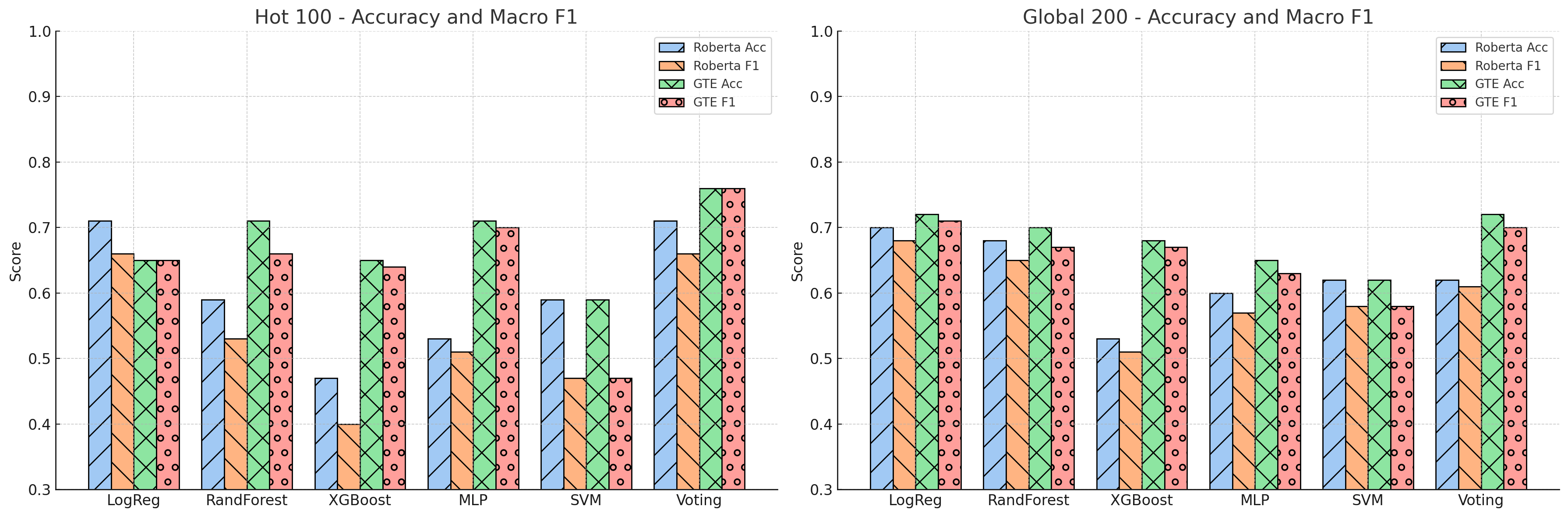}
        \caption{XLM-Roberta vs GTE Multilingual - Classification Metrics (Grouped Bars)}
        \label{fig:classification-results}
    \end{figure}

    \subsection{Classification Performance}
        
    \paragraph{Hot 100 songs:} Among songs that charted on the Billboard Hot 100, the best classification performance was achieved using \texttt{gte-multilingual-base} with a Voting Classifier, reaching an accuracy of \textbf{76\%} and a macro F1-score of \textbf{0.76}. With \texttt{xlm-roberta-base}, the strongest model was Logistic Regression, which achieved 71\% accuracy and a macro F1 of 0.66. These results indicate that the lyrics of Hot 100-charting songs contain learnable features that align with gendered performance styles. Interestingly, simpler models like Logistic Regression and ensemble methods outperformed more complex ones such as XGBoost or neural MLPs. This suggests that the distinction between male- and female-performed lyrics in the Hot 100 is linearly separable in the combined embedding-feature space. Handcrafted features such as \verb|kor_ratio|, \verb|filler_word_count|, and \verb|repetition_count| likely contributed to this separability, supported by semantic cues encoded in multilingual embeddings.
    
    \paragraph{Global 200 songs:} For the broader Global 200 chart, results followed a similar pattern. The best performance was obtained using \texttt{gte-multilingual-base} with Logistic Regression, achieving \textbf{72\% accuracy} and a \textbf{macro F1-score of 0.71}. \texttt{xlm-roberta-base} also yielded strong results with Logistic Regression (70\% accuracy, F1 0.68). Notably, models trained on Global 200 songs showed more balanced precision and recall across gender classes, compared to the Hot 100. This may reflect greater stylistic diversity or less polarisation in language usage across performers on the Global chart, possibly because it includes songs that are more locally or globally contextualised.
    
    Across both Billboard charts, results suggest that K-pop lyrics encode latent gendered markers that are detectable through multilingual sentence embeddings and lightweight classifiers. Although individual linguistic features (such as the Korean lyric ratio) did not yield significant differences across genders, their combination with learned representations allowed models to predict performer gender at a level significantly above chance. This indicates that lyrics used in Billboard-charting K-pop songs—especially those in the Hot 100—reflect stylised branding choices and that these choices manifest through language composition (Korean vs. English), syntactic structures (repetitions, filler words), and possibly thematic framing, all of which contribute to a latent performative identity that could potentially be algorithmically recovered.
    
    \begin{figure}[h!]
        \centering
        \includegraphics[width=0.75\linewidth]{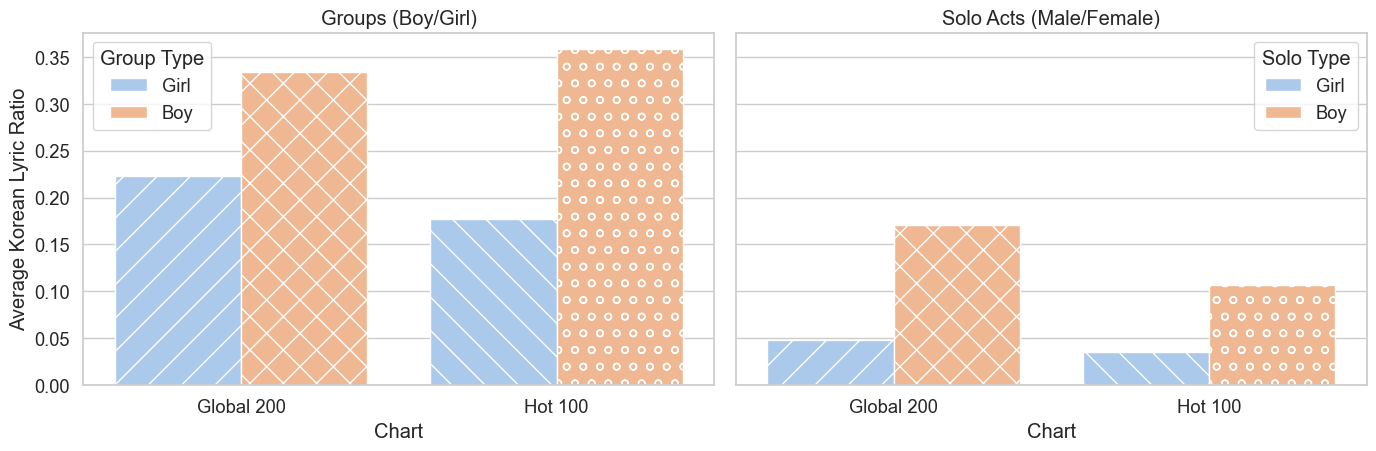}
        \caption{Average Korean Lyric Ratio by Group Type and Chart Types}
        \label{fig:stat-tests-1}
    \end{figure}
    
    \subsection{Code-Switching and Language Use Differences}

    The statistical tests are now examined to answer the research questions about English vs. Korean usage in charting K-pop songs. Figure~\ref{fig:stat-tests-1} illustrates the average Korean lyric ratio by group type, and we detail the statistical test results for both the Billboard Hot 100 and Global 200 charts below.
    
    \paragraph{English Code-Switching Frequency (Boy vs. Girl Groups):} Boy groups and girl groups exhibit remarkably similar code-switching frequencies across both charts, particularly on the Billboard Hot 100. On average, boy-group songs had 19.05 code-switches per song, while girl-group songs had 16.96 code-switches per song. A Welch’s t-test showed no significant difference ($t = 0.41$, $p = 0.686$). The distributions were non-normal (Shapiro-Wilk $p < 0.001$), and the Mann-Whitney U test likewise found no significant difference ($U = 555.5$, $p = 0.611$). On the Global 200 chart, the pattern was reversed—girl-group songs averaged slightly more switches (25.32) than boy-group songs (20.57)—but again, the difference was not statistically significant ($t = -1.20$, $p = 0.231$; Mann-Whitney $p = 0.634$). Overall, these findings confirm that \textbf{frequent language switching is a stylistic hallmark} across all K-pop groups, irrespective of gender, and is a consistent feature among Billboard-charting songs.
    
    \paragraph{Proportion of Korean vs. English (Boy vs. Girl Groups):} When analysing the proportion of Korean lyrics, a subtle gendered trend emerges across both charts. On the Hot 100, boy-group songs had a higher average Korean lyric ratio (0.24) compared to girl-group songs (0.11). Although Welch’s t-test suggested a significant difference ($p = 0.0097$), the non-normal distribution of the data made the Mann-Whitney U test more appropriate, which yielded a non-significant result ($p = 0.0796$). A similar pattern was observed on the Global 200, where boy-group songs again had a higher Korean ratio (0.24) than girl-group songs (0.17), though neither the Welch’s t-test ($p = 0.056$) nor the Mann-Whitney U test ($p = 0.105$) confirmed statistical significance. These findings point to a consistent, albeit non-significant, tendency for female-led acts to produce more English-dominant songs—possibly a strategic choice aimed at maximising global appeal, while male groups retain slightly more Korean content.
    
    \paragraph{Solo Artists – Gender Differences:} Our analysis also included solo acts, where the gender trend becomes more pronounced. Female soloists such as Lisa (“Money”) and Rosé (“On the Ground”) tended to release songs with English lyrics, often up to 95\% or more of the total. Male solo artists, by contrast, showed slightly more Korean retention, averaging around 10–17\% Korean. While sample sizes were small, the directional difference supports anecdotal claims~\cite{Auh2024Creativity, Garza2021Where} that female soloists in K-pop often craft tracks closely aligned with Western pop conventions, while male soloists (especially those from hip-hop or R\&B backgrounds) tend to maintain some Korean-language elements for authenticity or lyrical style.
    
    \paragraph{Chart Context (Hot 100 vs. Global 200):} To evaluate the impact of chart context on language use, we compared Korean lyric ratios for songs on the Hot 100 versus the Global 200. For boy groups, there was virtually no difference: the average Korean ratio was 0.236 on the Hot 100 and 0.238 on the Global 200, with both parametric and non-parametric tests confirming no significant difference ($p = 0.960$ and $p = 0.872$, respectively). In contrast, girl groups showed a modest decline in Korean content on the Hot 100 (0.112) compared to the Global 200 (0.171). While Welch’s t-test indicated borderline significance ($p = 0.0498$), the Mann-Whitney U test did not support this result ($p = 0.200$). Thus, no statistically significant difference can be confirmed, but the observed pattern suggests girl groups may lean toward English-heavy lyrics for U.S. chart success, whereas boy groups maintain consistent language use across charts, likely benefiting from strong international fan bases or genre-specific appeal.

    In conclusion, our analysis finds \textbf{no significant gender gap} in code-switching frequency across either chart, and only marginal trends suggesting boy groups retain slightly more Korean than girl groups. Both male and female K-pop acts heavily code-switch and predominantly use English in internationally successful tracks. Classification results further confirm the presence of stylistic or lexical distinctions, even if not solely attributable to the language ratio. This aligns with broader cultural observations that K-pop’s global rise has brought increasing English integration, with subtle variations based on gender, genre, and market orientation.

    \subsection{Error Analysis and Additional Observations}

    Several qualitative patterns emerged from misclassifications in the gender classification task (see Appendix~\ref{app:misclassification}). Songs that deviated from an artist’s typical style—such as collaborations or genre shifts—often confused the model. For instance, BLACKPINK tracks with prominent rap sections were sometimes misclassified as boy-group songs, while English-heavy BTS tracks like “Dynamite” were occasionally misclassified as girl-group songs, likely due to their pop style and high English content. Bigrams also varied across categories: girl-group songs featured English-heavy phrases like “love me” or “pink venom,” while boy-group lyrics included more Korean-English blends like “fake love” and rhythmic fillers like “yeah, yeah.” Solo tracks had distinctive hooks as well, reflecting individual artist branding. These stylistic and lexical tendencies likely contributed to model predictions and helped explain observed classification boundaries.

    \section{Conclusion}

    This study explored how code-switching and English usage in K-pop lyrics relate to global chart performance, with a particular focus on gendered patterns. English emerged as a dominant component in internationally charting K-pop songs, often constituting more than half of the lyrics for both boy and girl groups. Statistical tests showed no significant gender difference in code-switching frequency, indicating that alternating between Korean and English is a widespread stylistic norm. While male groups included slightly more Korean content on average, the difference lacked statistical significance. Female soloists, in contrast, frequently released songs with lyrics that were almost entirely in English, suggesting a stronger orientation toward global accessibility. Despite the absence of strong feature-level divergence, classification models trained on multilingual embeddings and linguistic features predicted performer gender with up to 76\% accuracy, confirming the presence of consistent stylistic cues. This pattern was most evident in Hot 100 charting songs, where female-group tracks leaned more heavily toward English than those on the Global 200, which featured comparatively more Korean content. Male groups, by contrast, showed little variation in language use across charts, potentially reflecting the influence of stable fan bases or genre. Overall, the findings underscore Code Switching with English's central role in K-pop's international appeal, while revealing subtle but detectable gendered tendencies in lyrical style.

    \section*{Ethics Statement: }
    
    This study did not involve the collection or use of any personal, identifiable, or private information relating to individuals. All data used consists of publicly available song lyrics and chart metadata. No social media posts, user accounts, or personally attributable content were accessed. As such, this research does not raise concerns related to human subjects or privacy, and no ethical review was required.
    
    \section*{Limitations and Future works: } 

    This study focused exclusively on K-pop songs that appeared on Billboard charts (Hot 100 and Global 200), which inherently biases the dataset toward internationally successful tracks. As a result, our findings may not generalise to the broader K-pop landscape, including domestic hits or songs by emerging artists. Additionally, the sample size—especially for statistical comparisons—remains modest, limiting our ability to detect smaller effects with high confidence. Some observed trends (e.g., linguistic differences between male and female artists) did not reach statistical significance and should be interpreted with caution.
    
    Importantly, our intention is not to draw gendered conclusions about K-pop performers or reinforce stereotypes. Rather, we treat male and female-led songs as distinct experimental groups to explore whether stylistic or linguistic differences exist, something suggested by the success of our classification models. The ability to predict group gender from lyrics with reasonable accuracy supports the premise that there are systematic differences worth examining, not as a matter of judgment but of stylistic characterisation. We also acknowledge that our classification methods simplify lyrics as feature vectors without engaging deeply with their cultural or semantic meaning. Future work could adopt a more nuanced linguistic or discourse-based approach, potentially in collaboration with Korean linguists to better interpret the functional and cultural dimensions of code-switching.
    
    There are several promising directions for future research. Expanding the dataset to include a broader range of K-pop eras and less globally prominent artists would allow for longitudinal analyses of how code-switching and English usage have evolved. Integrating audio features and visual aesthetics (e.g., music video analysis) could reveal correlations between language choice and genre or image. Another exciting avenue is to explore the predictive potential of lyrics and metadata (e.g., group type, language ratio, release context) in forecasting a song’s chart performance or peak position. Finally, studying listener reception—such as how English-heavy vs. Korean-heavy lyrics perform across different platforms or countries—could offer additional insights into the strategic use of code-switching in global music markets.
    

    \bibliography{ref}
    \appendix 
    
    \section{Appendix}


    \begin{table}[ht]
        \centering
        \caption{Hot 100 K-pop Artists by Category, Gender, and Number of Songs (Sorted)}
        \begin{tabular}{|p{3.5cm}|c|c|c|}
            \hline
            \textbf{Artist} & \textbf{Category} & \textbf{Gender} & \textbf{Num\_Songs} \\
            \hline
            BTS & G & M & \textbf{17} \\
            BLACKPINK & G & F & \textbf{6} \\
            V & S & M & \textbf{6} \\
            NewJeans & G & F & 5 \\
            Jennie & S & F & 5 \\
            Lisa & S & F & 5 \\
            J-Hope & S & M & 4 \\
            Rosé & S & F & 3 \\
            Jimin & S & M & 3 \\
            Jung Kook & S & M & 3 \\
            LE SSERAFIM & G & F & 2 \\
            TWICE & G & F & 2 \\
            Stray Kids & G & M & 2 \\
            Jin & S & M & 2 \\
            \hline
        \end{tabular}
    \end{table}
    
    \begin{table}[ht]
        \centering
        \caption{Global 200 K-pop Artists by Category, Gender, and Number of Songs (Sorted)}
        \begin{tabular}{|p{3.5cm}|c|c|c|}
            \hline
            \textbf{Artist} & \textbf{Category} & \textbf{Gender} & \textbf{Num\_Songs} \\
            \hline
            BLACKPINK & G & F & \textbf{18} \\
            NewJeans & G & F & \textbf{14} \\
            Jung Kook & S & M & \textbf{11} \\
            aespa & G & F & 11 \\
            Rosé & S & F & 10 \\
            Jennie & S & F & 9 \\
            Jimin & S & M & 9 \\
            V & S & M & 9 \\
            TWICE & G & F & 9 \\
            LE SSERAFIM & G & F & 7 \\
            Stray Kids & G & M & 7 \\
            Lisa & S & F & 6 \\
            SEVENTEEN & G & M & 6 \\
            BTS & G & M & 12 \\
            ITZY & G & F & 3 \\
            ENHYPEN & G & M & 3 \\
            J-Hope & S & M & 4 \\
            Jin & S & M & 4 \\
            RM & S & M & 3 \\
            Jisoo & S & F & 2 \\
            Red Velvet & G & F & 2 \\
            \hline
        \end{tabular}
    \end{table}

    \begin{CJK}{UTF8}{mj}
    
    \paragraph{Most Common Bigrams in Hot 100 Lyrics:}\label{app:misclassification}
    
    \begin{itemize}
        \item \textbf{Group – Female:} \textit{I know}, \textit{on my}, \textit{I, boy,}, \textit{I got}, \textit{it, whip}, \textit{All the}, \textit{you like}, \textit{Moonlight sunrise}, \textit{make you}
        \item \textbf{Group – Male:} \textit{Oh, oh,}, \textit{make it}, \textit{bring the}, \textit{in the}, \textit{I can}, \textit{fake love,}, \textit{love, fake}, \textit{me now}, \textit{yeah, yeah,}
        \item \textbf{Solo – Female:} \textit{아파트, 아파트}, \textit{ladies run}, \textit{I'm a}, \textit{JENNIE, JENNIE,}, \textit{on the}, \textit{아파트 아파트,}, \textit{be born}, \textit{my ladies}, \textit{call me}, \textit{the ground}
        \item \textbf{Solo – Male:} \textit{go, go,}, \textit{Chicken noodle}, \textit{noodle soup}, \textit{I like}, \textit{Never let}, \textit{love me}, \textit{next to}, \textit{let go,}, \textit{soup Chicken}
    \end{itemize}
    
    \vspace{0.5cm}
    
    \paragraph{Most Common Bigrams in Global 200 Lyrics:}
    
    \begin{itemize}
        \item \textbf{Group – Female:} \textit{tok, tik,}, \textit{tik, tok,}, \textit{tik, tik,}, \textit{I don't}, \textit{with my}
        \item \textbf{Group – Male:} \textit{get it}, \textit{hot, hot,}, \textit{it done,}, \textit{I wanna}, \textit{you know}, \textit{hot, hot}
        \item \textbf{Solo – Female:} \textit{break, break,}, \textit{to be}, \textit{I did}, \textit{I need}, \textit{Damn, right,}, \textit{I just}
        \item \textbf{Solo – Male:} \textit{to be}, \textit{I want}, \textit{let you}, \textit{you to}, \textit{want you}
    \end{itemize}

    \end{CJK}
    
    \subsection{Statistical Tests in detail}
    
    \begin{table}[ht!]
    \centering
    \begin{adjustbox}{max width=\textwidth}
    \begin{tabular}{@{}p{3.5cm}p{4cm}ccccc@{}}
    \toprule
    \textbf{Feature} & \textbf{Comparison} & \textbf{Means} & \textbf{t-test (p)} & \textbf{U-test (p)} & \textbf{Normality} & \textbf{Conclusion} \\
    \midrule
    Code-switching & Boy groups vs. girl groups (Hot 100) & 19.05 vs. 16.96 & 0.6862 & 0.6109 & Not normal & Not significant \\
    Korean lyric ratio & Boy groups vs. girl groups (Hot 100) & 0.24 vs. 0.11 & \textbf{0.0097} & 0.0796 & Not normal & Not significant \\
    \addlinespace
    Code-switching & Boy groups vs. girl groups (Global 200) & 20.57 vs. 25.32 & \textbf{0.2312} & 0.6341 & Not normal & Not significant \\
    Korean lyric ratio & Boy groups vs. girl groups (Global 200) & 0.24 vs. 0.17 & 0.0558 & 0.1054 & Not normal & Not significant \\
    \addlinespace
    Korean lyric ratio & Boy groups: Hot 100 vs. Global 200 & 0.236 vs. 0.238 & 0.9599 & 0.8717 & Not normal & Not significant \\
    Korean lyric ratio & Girl groups: Hot 100 vs. Global 200 & 0.112 vs. 0.171 & \textbf{0.0498} & 0.1995 & Not normal & Not significant \\
    \bottomrule
    \end{tabular}
    \end{adjustbox}
    \caption{Summary of statistical test results comparing code-switching and Korean lyric ratio across gender and chart conditions.}
    \label{tab:stat-tests-complete}
    \end{table}
    
\end{document}